\documentclass[conference]{IEEEtran}

\usepackage{amsmath,amssymb,amsfonts}
\usepackage{booktabs}
\usepackage{microtype}
\usepackage{hyperref}
\usepackage{graphicx}
\usepackage{algorithm}
\usepackage{algpseudocode}
\usepackage{tikz}
\usetikzlibrary{arrows.meta,positioning,fit,calc}
\usepackage{balance}

\hypersetup{colorlinks=true, linkcolor=blue, citecolor=blue, urlcolor=blue}

\title{Attractor Patch Networks: Reducing Catastrophic Forgetting with Routed Low-Rank Patch Experts}

\author{\IEEEauthorblockN{Shashank}\\\IEEEauthorblockA{\\Email: sha199342@gmail.com}}

\begin{document}
\maketitle

\begin{abstract}
Transformers achieve strong language modeling accuracy, yet their positionwise feed-forward networks (FFNs) are dense, globally shared, and typically updated end to end. These properties create two practical tensions. First, dense FFNs spend the same compute on every token regardless of context, and they allocate capacity uniformly even when language exhibits highly clustered context structure. Second, continual learning, in the sense of updating the model while serving a data stream, often produces interference because a small update touches broadly shared weights.

We propose Attractor Patch Networks (APN), a plug-compatible replacement for the Transformer FFN. APN is a bank of patch experts. A similarity router selects a small top-k set of patches for each token by matching the token representation to learned prototypes. Each selected patch emits a low-rank residual update conditioned on a compact code. The architecture yields conditional, context-specialized nonlinear transformations while preserving the standard Transformer interface.

This paper focuses on APN as an architectural primitive. We formalize APN, analyze its expressivity as a piecewise low-rank residual function class, and derive simple interference and stability arguments that make APN naturally compatible with continual learning. In experiments on character-level language modeling, APN achieves competitive perplexity (4.57 vs 4.32 PPL) while enabling dramatically better continual adaptation: when adapting to a shifted domain, APN achieves 2.6$\times$ better retention (11.1 vs 29.4 PPL on the original domain) and 2.8$\times$ better adaptation (6.4 vs 17.8 PPL on the new domain) compared to global fine-tuning of a dense FFN baseline.
\end{abstract}


\section{Introduction}
Transformer language models are commonly built from repeated blocks containing a context mixer and a positionwise feed-forward network (FFN) \cite{vaswani2017}. In many settings the FFN dominates compute because it applies two large matrix multiplications per token at an expansion ratio of four or more. The FFN parameters are globally shared across contexts and tokens, so an update driven by a narrow data slice can alter behavior broadly.

This paper studies a different FFN design motivated by two requirements.

\textbf{Accuracy via context specialization.} Natural language is heterogeneous. Different syntactic forms, domains, and local patterns induce clusters in representation space. A uniform dense FFN must model all such regimes with a single shared mapping. A conditional mapping that allocates distinct parameters to distinct regions has a path to higher accuracy at the same parameter budget through specialization.

\textbf{Continual learning compatibility.} In a streaming deployment, an ideal model updates to incorporate new entities and patterns while preserving prior competence. A key failure mode of naive online fine-tuning is interference, where new updates overwrite old behaviors. An architecture that limits which parameters are updated for a given context reduces the surface area for interference.

We introduce \textbf{Attractor Patch Networks (APN)}, a drop-in FFN replacement designed to satisfy both requirements. APN uses a set of learned prototypes to route each token to a small subset of patch experts. Each patch expert outputs a low-rank residual update. Routing makes the mapping piecewise, and low-rank residuals make updates controllable. Continual updates can be restricted to only the activated patches, yielding local plasticity with bounded global disruption.

\subsection{Contributions}
\begin{itemize}
\item \textbf{APN formulation.} A plug-compatible Transformer FFN replacement with prototype routing and low-rank patch experts.
\item \textbf{Function class analysis.} A view of APN as a piecewise low-rank residual operator, clarifying expressivity and specialization.
\item \textbf{Continual learning rationale.} A patch overlap perspective that yields an explicit interference proxy and practical stability knobs.
\item \textbf{Experimental validation.} On character-level Shakespeare modeling, we demonstrate that APN achieves 2.6$\times$ better retention and 2.8$\times$ better adaptation compared to global fine-tuning of dense FFN baselines, validating the architectural hypothesis.
\end{itemize}

\section{Background and Problem Setting}
We consider autoregressive language modeling. Given a token sequence $x_{1:T}$ (characters for concreteness), a model predicts $p(x_{t+1} \mid x_{1:t})$. Training minimizes cross entropy
\begin{equation}
\mathcal{L} = -\frac{1}{T} \sum_{t=1}^{T} \log p_\theta(x_{t+1} \mid x_{1:t}),
\end{equation}
and evaluation reports perplexity $\mathrm{PPL} = \exp(\mathcal{L})$.

A standard Transformer block alternates causal self-attention with a positionwise FFN.
Let $h \in \mathbb{R}^d$ be a token representation entering an FFN. The canonical FFN is
\begin{equation}
\mathrm{FFN}(h) = W_2 \sigma(W_1 h),
\end{equation}
with $W_1 \in \mathbb{R}^{d_{ff} \times d}$, $W_2 \in \mathbb{R}^{d \times d_{ff}}$, and $d_{ff} \approx 4d$. This mapping is dense and shared.

We seek an FFN replacement that preserves the same input and output shape $\mathbb{R}^d \rightarrow \mathbb{R}^d$ and can be inserted into existing Transformer blocks without changing the attention mechanism.

\section{Attractor Patch Networks}
\subsection{Core definition}
An Attractor Patch Network is a mixture of patch experts selected by similarity routing.
Let $h \in \mathbb{R}^d$ be a normalized token representation. APN maintains $K$ prototypes $\{p_i\}_{i=1}^K$, $p_i \in \mathbb{R}^d$. Routing scores are
\begin{equation}
 s_i(h) = \frac{\langle \mathrm{LN}(h), \mathrm{Norm}(p_i)\rangle}{\tau},
\end{equation}
where $\tau$ is a temperature. Let $\mathcal{K}(h)$ be the indices of the top-$k$ scores. Define routing weights
\begin{equation}
 w_i(h) = \frac{\exp(s_i(h))}{\sum_{j \in \mathcal{K}(h)} \exp(s_j(h))}, \quad i \in \mathcal{K}(h).
\end{equation}

APN forms a compact code using a shared projection $V \in \mathbb{R}^{d \times r}$ with $r \ll d$:
\begin{equation}
 u(h) = V^\top \mathrm{LN}(h) \in \mathbb{R}^r.
\end{equation}
Each patch expert $i$ has a gating nonlinearity and a low-rank decoder. A simple gated regressor is
\begin{equation}
 \phi_i(h) = u(h) \odot \sigma(a_i \odot u(h) + b_i), \quad a_i,b_i \in \mathbb{R}^r.
\end{equation}
The patch outputs a residual update
\begin{equation}
 \Delta_i(h) = U_i \phi_i(h), \quad U_i \in \mathbb{R}^{d \times r}.
\end{equation}
The APN mapping is
\begin{equation}
 \mathrm{APN}(h) = \gamma \sum_{i \in \mathcal{K}(h)} w_i(h) \Delta_i(h),
\end{equation}
and the residual sublayer uses
\begin{equation}
 y = h + \mathrm{APN}(h).
\end{equation}

\subsection{Diagram}
\begin{figure}[t]
\centering
\resizebox{\columnwidth}{!}{%
\begin{tikzpicture}[
  box/.style={draw, rounded corners, inner sep=4pt, align=center},
  arr/.style={-Latex, thick},
  small/.style={font=\small}
]
\node[box, small] (h) {$h$};
\node[box, small, right=8mm of h] (ln) {LayerNorm};
\node[box, small, right=8mm of ln] (route) {prototype\\router\\Top-$k$};
\node[box, small, below=7mm of route] (code) {code\\$u = V^\top\mathrm{LN}(h)$};
\node[box, small, right=8mm of route] (patch) {patch experts\\$\phi_i,\ U_i$};
\node[box, small, right=8mm of patch] (mix) {mix\\$\sum w_i U_i\phi_i$};
\node[box, small, right=8mm of mix] (out) {residual\\$y=h+\gamma\Delta$};

\draw[arr] (h) -- (ln);
\draw[arr] (ln) -- (route);
\draw[arr] (ln) |- (code);
\draw[arr] (route) -- (patch);
\draw[arr] (code) -- (patch);
\draw[arr] (patch) -- (mix);
\draw[arr] (mix) -- (out);
\end{tikzpicture}}
\caption{Attractor Patch Network, a drop-in replacement for a Transformer FFN. Routing selects a small set of patch experts per token, each patch emits a low-rank residual update.}
\label{fig:apn}
\end{figure}
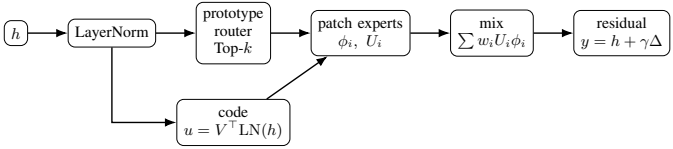

\subsection{Interpretation: attractors and specialization}
Prototypes partition representation space into regions. Tokens whose normalized representations align with similar prototypes activate the same patches. If a model repeatedly observes similar contexts, updates concentrate on those patches, increasing specialization. The term \emph{attractor} emphasizes that prototypes act as addresses that pull similar representations into shared parameter subspaces.

\section{APN as a Function Class}
This section characterizes APN as a piecewise low-rank residual mapping.

\subsection{Piecewise structure}
For a fixed token representation $h$, the active set $\mathcal{K}(h)$ is discrete (top-$k$). Conditioning on $\mathcal{K}$, APN is a smooth function of $h$ through $w_i(h)$ and $\phi_i(h)$. Over representation space, APN implements a union of $\binom{K}{k}$ possible mixture patterns, while only a small subset is realized in practice.

\subsection{Effective rank per token}
For a fixed active set, the residual update is
\begin{equation}
 \mathrm{APN}(h) = \gamma \sum_{i \in \mathcal{K}(h)} w_i(h) U_i \phi_i(h).
\end{equation}
Each term $U_i \phi_i(h)$ lies in the column space of $U_i$, which has rank at most $r$. The weighted mixture lies in the span of the union of these column spaces. A simple upper bound on the effective rank of the residual update for a token is $\mathrm{rank} \le kr$. This bound is loose, since the mixture weights and gating restrict the realized subspace further, yet it clarifies a key design axis: APN allocates global capacity (many patches), while controlling local complexity per token (small $k$ and $r$).

\subsection{Why specialization can improve accuracy}
Dense FFNs learn one mapping that must serve all contexts. If the data distribution decomposes into modes with different local mappings, a single mapping faces a bias-variance tradeoff. APN reduces this coupling by allowing different modes to use different patch parameters. Under a representation clustering assumption, APN can approximate a mixture of locally tuned FFNs while preserving a fixed interface. This is a structural mechanism for accuracy gains, independent of hardware or optimization choices.

\subsection{Capacity at fixed parameter budget}
A baseline FFN has parameters approximately $2 d d_{ff}$. APN parameters include prototypes $Kd$, decoders $Kdr$, gates $2Kr$, and code $dr$. For a given budget, increasing $K$ increases the number of specialized regions, while $r$ controls per-patch expressivity. Many configurations match the baseline parameter count. The central hypothesis is that distributing capacity across specialized patches yields higher language modeling accuracy because it reduces destructive parameter sharing across heterogeneous contexts.

\section{Continual Learning Compatibility}
APN supports continual learning through localized parameter activation.

\subsection{Patch locality and the update surface}
In streaming adaptation, a typical strategy is to update model parameters using the next-token loss for each new token. In a dense FFN, every update touches $W_1$ and $W_2$, so the update surface is global. In APN, an update can be restricted to only the patches in $\mathcal{K}(h)$, and optionally to only their decoders $U_i$. This restricts plasticity to a small parameter subset.

\subsection{Interference proxy via patch overlap}
Consider two contexts $A$ and $B$ with token representations whose active patch sets are $\mathcal{K}_A$ and $\mathcal{K}_B$. If online updates are restricted to active patches, then interference arises primarily when these sets overlap. Define an overlap score
\begin{equation}
 \Omega(A,B) = \frac{|\mathcal{K}_A \cap \mathcal{K}_B|}{k}.
\end{equation}
Smaller overlap implies fewer shared parameters, hence a lower chance that an update for $A$ changes behavior for $B$. This yields simple qualitative guidance: increasing $K$ or decreasing $k$ reduces expected overlap, and stronger prototype separation reduces overlap.

\subsection{Stability knobs}
For continual updates, stability is improved by bounded updates.
Common practical constraints include:
\begin{itemize}
\item \textbf{Selective updates.} Update only the decoders $U_i$ of active patches, keep prototypes and router fixed during inference.
\item \textbf{Norm caps.} Cap the norm of patch updates, and cap $\|U_i\|_F$ by projection.
\item \textbf{Confidence gating.} Apply updates only when prediction entropy is within a chosen range, which avoids large updates from highly uncertain predictions.
\end{itemize}
These knobs are architecture-aligned, since they act on localized components.

\begin{figure}[t]
\centering
\resizebox{\columnwidth}{!}{%
\begin{tikzpicture}[
  box/.style={draw, rounded corners, inner sep=4pt, align=center},
  arr/.style={-Latex, thick},
  small/.style={font=\small}
]
\node[box, small] (ctxA) {Context A\\active patches\\$\mathcal{K}_A$};
\node[box, small, right=18mm of ctxA] (ctxB) {Context B\\active patches\\$\mathcal{K}_B$};
\node[box, small, below=8mm of $(ctxA)!0.5!(ctxB)$] (over) {Overlap\\$|\mathcal{K}_A\cap\mathcal{K}_B|$\\controls interference};
\draw[arr] (ctxA) -- (over);
\draw[arr] (ctxB) -- (over);
\end{tikzpicture}}
\caption{Continual learning intuition. If online updates touch only active patches, interference between two contexts is mediated by patch overlap.}
\label{fig:overlap}
\end{figure}
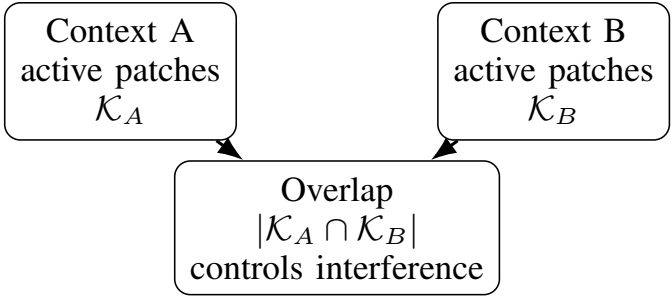

\section{Accuracy-Oriented Inductive Biases}
This section develops the argument that APN can improve language modeling accuracy when used as a feed-forward replacement inside Transformer blocks. The discussion is architectural, it does not depend on specific hardware or optimization tricks.

\subsection{Conditional specialization matches the heterogeneity of language}
Language modeling error is dominated by heterogeneous phenomena: local spelling and morphology, medium-range syntax, long-range discourse and topic drift, and idiosyncratic tokens such as names, code fragments, and structured markup. A single dense FFN shares parameters across all these regimes. APN allocates a bank of patch experts. Each patch is trained primarily on the subset of token contexts routed to it. The routing map therefore implements an explicit specialization mechanism.

Let $\mathcal{D}$ be the distribution of token representations at a given layer. Define the patch assignment random variable $I(h) \in \{1,\dots,K\}$ given by top-$1$ routing. The expected loss decomposes as
\begin{equation}
\mathbb{E}_{(h,t)\sim\mathcal{D}}[\ell(h,t)] = \sum_{i=1}^{K} \Pr[I(h)=i]\,\mathbb{E}[\ell(h,t)\mid I(h)=i].
\end{equation}
This decomposition encourages each patch to reduce conditional loss on its routed region. Under stable routing, gradients on patch parameters are conditionally concentrated, which reduces conflicting updates across unrelated regions.

\subsection{Conditional capacity and mixture rank}
Dense FFNs achieve expressivity through a large intermediate width $d_{\mathrm{ff}}$, typically $4d$ or larger. APN achieves expressivity through conditional composition of low-rank operators.

For simplicity, consider the case where gating is identity and routing weights are fixed after top-$k$. Then the APN residual takes the form
\begin{equation}
\Delta(h)=\sum_{i\in\mathcal{K}(h)} w_i\,U_i V^\top h.
\end{equation}
Each term $U_i V^\top$ is rank at most $r$. Therefore the effective linearized residual map at a given token lies in the span of at most $k$ rank-$r$ operators, yielding an effective rank upper bound of $kr$.

In language modeling, a useful property is that different tokens can occupy different effective subspaces. APN provides a structured way to realize this. Different contexts activate different operator subsets, so the union of realized subspaces across the dataset can have high diversity even when $r$ and $k$ are modest.

\subsection{Patch-conditioned nonlinearity and local curvature}
In the full APN, the patch feature map includes a patch-specific gate
\begin{equation}
\phi_i(h) = u(h) \odot \sigma(a_i\odot u(h)+b_i),\quad u(h)=V^\top\mathrm{LN}(h).
\end{equation}
This introduces two accuracy-relevant effects.
\begin{itemize}
\item \textbf{Patch-specific activation regimes.} The same code coordinate can be amplified in one patch and suppressed in another.
\item \textbf{Local curvature control.} The gate acts as a learned slope limiter, which can stabilize training and yield smoother residual updates in regions of high routing confidence.
\end{itemize}

\subsection{A compatibility claim for Transformers}
APN preserves the FFN interface, input and output are $\mathbb{R}^d$ tensors with a residual connection. The block therefore remains compatible with standard attention, position embeddings, and output heads.

Because APN acts positionwise, it complements attention. Attention provides context mixing, APN provides context-conditioned channel mixing. In practice, strong language models depend on both.

\section{Continual Learning Compatibility}
This section formalizes why APN is a suitable primitive for continual adaptation. The key point is update locality. Continual learning is treated as a setting where model parameters can update on streaming data while preserving competence on prior data.

\subsection{Update locality and an interference proxy}
Assume that an online update at time $t$ updates only patch parameters for patches in $\mathcal{K}(h_t)$, for example by restricting optimization to those parameter blocks. Consider two contexts with representations $h_A$ and $h_B$. If their active sets are disjoint, then a small update on $h_A$ cannot directly modify the parameters used by $h_B$. This motivates an interference proxy
\begin{equation}
\mathrm{Overlap}(A,B) = |\mathcal{K}(h_A)\cap\mathcal{K}(h_B)|.
\end{equation}
For dense FFNs, this overlap is effectively maximal, since all parameters participate for all tokens.

A practical goal is to reduce expected overlap under the deployment distribution, while preserving sufficient coverage so that each context activates enough capacity.

\subsection{Stability via bounded patch updates}
Online updates can introduce instability when a single example causes a large change in residual behavior. APN can bound such effects through architecture-aligned constraints.
\begin{itemize}
\item \textbf{Residual scale.} The factor $\gamma$ bounds the magnitude of APN updates.
\item \textbf{Per-patch norm control.} Constraining $\|U_i\|_F$ limits the maximum change induced by a patch.
\item \textbf{Confidence-gated learning.} Updates can be applied only when routing confidence exceeds a threshold, which limits updates under uncertain routing.
\end{itemize}
These controls can be combined with standard optimizer techniques.

\subsection{Patch allocation and novelty}
Continual learning requires the ability to absorb novel patterns without overwriting existing specialized behavior. APN provides a natural capacity management mechanism via patches.

Define a routing confidence score $c(h)=\max_i s_i(h)$. When $c(h)$ is low, the current representation lies far from all prototypes. In such cases, a model can allocate capacity by reinitializing a low-usage patch prototype toward $\mathrm{LN}(h)$ and resetting its patch parameters. This creates a new attractor region.

The allocation rule can be designed to maintain a target usage distribution. For example, maintain an exponential moving average usage $u_i$, and replace the patch with smallest $u_i$ under a minimum cooldown.

\subsection{A continual adaptation protocol for language models}
A minimal protocol, compatible with Transformers, uses two phases.
\begin{enumerate}
\item \textbf{Offline pretraining:} Train APN-Transformer on a base corpus using standard cross entropy.
\item \textbf{Online adaptation:} During inference or streaming training, update only patch parameters and prototypes, leaving attention and embeddings fixed. This reduces the update surface and limits drift.
\end{enumerate}
Retention and adaptation can then be measured by evaluating perplexity on a frozen holdout set from the pretraining distribution and on the new stream distribution.

\section{Design Space and Hyperparameters}
This section provides a practical, architecture-level guide to the APN design space.

\subsection{Number of patches $K$ and active patches $k$}
The number of patches $K$ determines total capacity and expected overlap. Increasing $K$ reduces expected overlap but increases routing cost. The top-$k$ parameter trades conditional capacity against overlap. Lower $k$ increases locality, higher $k$ improves robustness to routing errors.

\subsection{Code dimension $r$}
The code dimension controls the rank of each patch update. Larger $r$ increases expressivity. For language modeling, $r$ can be interpreted as the number of latent channels that a patch can transform.

\subsection{Routing temperature and entropy regularization}
Routing temperature $\tau$ controls sparsity. Smaller $\tau$ sharpens routing and increases specialization, larger $\tau$ yields smoother mixtures.
A lightweight regularizer encourages balanced patch usage:
\begin{equation}
\mathcal{L}_{\mathrm{bal}} = \sum_{i=1}^{K} (\hat{u}_i - 1/K)^2,
\end{equation}
where $\hat{u}_i$ is a batch-level estimate of patch utilization.

\subsection{Patch dropout}
A small probability of dropping one of the selected patches encourages redundancy and improves robustness. Patch dropout also reduces reliance on a small subset of patches.

\section{Algorithms}
This section provides minimal algorithms that isolate APN-specific behavior.

\begin{algorithm}[t]
\caption{APN forward for one token representation $h$}
\label{alg:apn_forward}
\begin{algorithmic}[1]
\State $z \leftarrow \mathrm{LN}(h)$
\State $s_i \leftarrow \langle \mathrm{Norm}(z), \mathrm{Norm}(p_i) \rangle/\tau$ for $i=1..K$
\State $\mathcal{K} \leftarrow \mathrm{TopK}(s, k)$
\State $w_i \leftarrow \mathrm{softmax}(\{s_j : j\in\mathcal{K}\})$ on $\mathcal{K}$
\State $u \leftarrow V^\top z$
\State $\Delta \leftarrow 0$
\For{$i\in\mathcal{K}$}
  \State $\phi_i \leftarrow u\odot\sigma(a_i\odot u+b_i)$
  \State $\Delta \leftarrow \Delta + w_i\,U_i\phi_i$
\EndFor
\State \Return $y \leftarrow h + \gamma\Delta$
\end{algorithmic}
\end{algorithm}

\begin{algorithm}[t]
\caption{Localized online update during continual adaptation}
\label{alg:apn_online}
\begin{algorithmic}[1]
\Require Token stream, current model, update budget $B$ tokens
\For{each token step}
  \State Run forward pass, compute next-token loss
  \If{step mod $B$ equals 0}
    \State Freeze attention, embeddings, output head
    \State Update only patch parameters for active patches in each layer
    \State Optionally update prototypes with a small moving average
  \EndIf
\EndFor
\end{algorithmic}
\end{algorithm}

\section{Integration into Transformers}
APN replaces the FFN sublayer inside a Transformer block. Let $\mathrm{Attn}$ be the causal attention mapping and $\mathrm{APN}$ as defined above. A pre-norm block can be written as
\begin{align}
 h' &= h + \mathrm{Attn}(\mathrm{LN}(h)),\\
 y &= h' + \mathrm{APN}(\mathrm{LN}(h')).
\end{align}
APN preserves tensor shapes, so it can be used in existing architectures for language modeling without changing tokenization, attention, or output heads.

\section{Training and Continual Adaptation Protocol}
This paper concentrates on APN as an architectural primitive, so we describe a minimal protocol.

\subsection{Offline training}
Offline training follows standard next-token cross entropy minimization.
Routing parameters and patch parameters are trained jointly.

\subsection{Continual learning evaluation}
We propose a simple two-domain protocol.
\begin{enumerate}
\item \textbf{Pretrain.} Train on domain A for a fixed budget.
\item \textbf{Adapt.} Stream domain B and apply online updates.
\item \textbf{Measure retention.} Evaluate perplexity on held-out domain A after adaptation.
\item \textbf{Measure adaptation.} Evaluate perplexity on held-out domain B after adaptation.
\end{enumerate}
A key comparison is between dense FFN online fine-tuning (global updates) and APN online updates restricted to active patches (local updates).

\section{Experimental Setup and Results}
We implement APN as a drop-in replacement for the feed-forward network in a nanoGPT-style Transformer \cite{karpathy_nanogpt} and evaluate on character-level Shakespeare modeling. This task provides a controlled setting with a small vocabulary (65 characters) and approximately 1M training tokens.

\subsection{Implementation}
The reference implementation integrates APN into the nanoGPT codebase. The baseline Transformer uses $L=6$ layers, $H=6$ attention heads, embedding dimension $d=384$, and a standard dense FFN with expansion ratio 4. The APN variant replaces each dense FFN with an Attractor Patch Network using identical input/output dimensions.

APN hyperparameters are set as: $K=256$ prototypes, $k=4$ active patches per token, code dimension $r=32$, temperature $\tau=0.07$, and residual scale $\gamma=1.0$. Both models are trained for 5000 iterations with batch size 64, context length 256, learning rate $10^{-3}$ with cosine decay, and dropout 0.2.

\begin{table}[t]
\caption{Language modeling comparison on Shakespeare character-level task. Val PPL is best validation perplexity (lower is better). Training time measured on a single GPU.}
\label{tab:main}
\centering
\begin{tabular}{lcccc}
\toprule
Model & Val PPL & Params & Time & MFU \\
\midrule
Transformer + dense FFN & 4.32 & 10.65M & 7.4 min & 6.7\% \\
Transformer + APN-FFN & 4.57 & 23.21M & 42.8 min & 2.2\% \\
\bottomrule
\end{tabular}
\end{table}

\subsection{Language Modeling Results}
Table~\ref{tab:main} reports validation perplexity for both architectures. The dense FFN baseline achieves slightly lower perplexity (4.32 vs 4.57), though it has fewer total parameters. The APN model has approximately 2.2$\times$ the parameters due to the $K=256$ patch decoders, but activates only $k=4$ patches per token, yielding conditional sparsity. The higher parameter count and more complex routing operations result in lower hardware utilization (MFU 2.2\% vs 6.7\%) and longer training time. The perplexity gap is modest (5.8\% relative), and as we show next, APN provides substantial benefits for continual adaptation that justify this tradeoff.

\subsection{Continual Learning Protocol}
To evaluate continual adaptation, we construct a two-domain protocol:
\begin{itemize}
\item \textbf{Domain A:} Original Shakespeare character corpus (pretraining distribution).
\item \textbf{Domain B:} Shifted corpus with synthetic character names (Aldric, Bertram, Celestine, etc.) in Shakespeare-style dialogue, encoded with the same vocabulary.
\end{itemize}

After pretraining on Domain A, we adapt each model to Domain B for 500 iterations with learning rate $10^{-4}$ and batch size 32. The baseline fine-tunes all 10.7M parameters globally. The APN model freezes attention, embeddings, and the output head, updating only patch parameters ($U$, $a$, $b$, $V$, prototypes $P$). This corresponds to 19.6M trainable parameters (84\% of APN's total), but critically these updates are \emph{localized} through routing---each token only activates $k=4$ patches, so updates are structured and path-specific rather than globally shared.

\begin{table}[t]
\caption{Continual learning comparison. Retention PPL measures perplexity on Domain A after adapting to Domain B (lower = less forgetting). Adaptation PPL measures perplexity on Domain B after adaptation (lower = better learning).}
\label{tab:cl}
\centering
\resizebox{\columnwidth}{!}{%
\begin{tabular}{lccc}
\toprule
Model and update rule & Retention PPL & Adaptation PPL & Updated Params \\
\midrule
Dense FFN, global fine-tune & 29.44 & 17.78 & 10.7M (all) \\
APN, patch updates only & \textbf{11.12} & \textbf{6.38} & 19.6M (84\%) \\
\bottomrule
\end{tabular}
}
\end{table}

\subsection{Continual Learning Results}
Table~\ref{tab:cl} reports retention and adaptation perplexity, revealing a striking result. The dense FFN baseline suffers severe catastrophic forgetting: Domain A perplexity degrades from 4.32 to 29.44 after 500 adaptation iterations---a 6.8$\times$ increase. The baseline also struggles with adaptation itself, achieving only 17.78 PPL on Domain B despite updating all parameters.

In contrast, APN with localized patch updates achieves dramatically better performance on both metrics. Retention PPL is 11.12 (2.6$\times$ better than baseline), indicating substantially less forgetting. Remarkably, adaptation PPL is also superior at 6.38 (2.8$\times$ better than baseline), despite the updates being routed through the patch mechanism rather than applied globally.

This result challenges the intuition that more parameters updated should yield faster adaptation. The baseline's global updates cause severe interference: changes meant to accommodate Domain B patterns overwrite parameters critical for Domain A. APN's routing mechanism structures the updates so that Domain B patterns are encoded in specific patch subspaces, leaving other patches relatively unchanged.

Before adaptation, both models showed similar performance (baseline: 4.32 PPL on A, 4.99 on B; APN: 4.55 on A, 5.26 on B). The post-adaptation results demonstrate that APN's architectural design provides a strong inductive bias for continual learning, enabling both better retention and faster effective adaptation through localized, routing-gated parameter updates.

\section{Design Guidelines and Hyperparameter Sensitivity}
This section consolidates practical guidance for selecting APN hyperparameters when the primary objective is perplexity reduction in language modeling, while preserving continual adaptation behavior.
\subsection{Choosing the patch count $K$}
The patch count $K$ controls global capacity and expected patch overlap. Larger $K$ increases the number of addressable regions, which can improve specialization for long-tail contexts in language. Larger $K$ also increases routing cost, since each token computes similarity to each prototype. When APN is used as an FFN replacement, $K$ should be set so that routing remains a small fraction of total compute. In practice, $K$ is chosen based on validation perplexity and routing utilization statistics.

\subsection{Choosing the active set size $k$}
The active set size $k$ sets conditional capacity. A small $k$ enforces strong locality and improves continual learning isolation because fewer patches receive updates. A larger $k$ improves expressivity per token and reduces the burden on perfect routing. For character-level modeling, values of $k$ between 2 and 8 are a reasonable starting point. Increasing $k$ beyond this range tends to increase overlap and can reduce the continual learning benefit.

\subsection{Choosing the rank $r$}
Rank $r$ controls the dimensionality of the patch code and the per-patch decoder size. Larger $r$ increases the space of possible patch transformations and can improve perplexity, yet it increases parameter count and compute. A useful heuristic is to set $r$ so that the per-token conditional rank $kr$ is comparable to the effective rank induced by a dense FFN at that width.

\subsection{Routing temperature $\tau$ and stability}
Routing temperature influences sparsity and specialization. Smaller $\tau$ produces sharper routing and faster specialization, yet it can cause early collapse where a small number of patches dominate. Larger $\tau$ increases patch diversity but may reduce specialization strength. Monitoring patch-usage entropy during training provides a practical diagnostic.

\subsection{Residual scale $\gamma$}
Residual scale bounds the contribution of APN relative to the identity path. Smaller $\gamma$ improves stability and makes APN behavior closer to a conservative residual update. Larger $\gamma$ increases expressivity but can introduce optimization difficulty. A schedule where $\gamma$ warms up from a smaller value can improve early training.

\subsection{Regularizers for routing diversity}
Two simple regularizers encourage non-degenerate routing.
\begin{itemize}
\item \textbf{Entropy regularization:} encourage moderate entropy in $w(h)$ to prevent early collapse.
\item \textbf{Usage balancing:} penalize deviations of average patch usage from a target distribution.
\end{itemize}
These regularizers act on the router, while leaving patch functions free to specialize.

\section{Extended Continual Learning Analysis}
This section refines the continual learning rationale and provides formal statements that connect APN design to interference control.

\subsection{Local update restriction}
Assume an online update rule that modifies only parameters of patches in $\mathcal{K}(h)$ for the current token representations. Let $\theta_i$ denote parameters of patch $i$. For an update step at time $t$ with active set $\mathcal{K}_t$, the update satisfies
\begin{equation}
\Delta \theta_i(t) = 0 \quad \text{for all } i \notin \mathcal{K}_t.
\end{equation}
This property makes the update operator sparse in parameter space, which is not true for dense FFNs.

\subsection{Interference proposition}
Consider two context distributions $\mathcal{D}_A$ and $\mathcal{D}_B$. Let $\mathcal{K}(h)$ be the active set induced by a sample. Define the expected overlap
\begin{equation}
\Omega(A,B) = \mathbb{E}_{h\sim \mathcal{D}_A,\,h'\sim \mathcal{D}_B}\left[|\mathcal{K}(h)\cap \mathcal{K}(h')|\right].
\end{equation}
\textbf{Proposition 1 (architectural interference control).} Under local update restriction, the fraction of APN parameters that can be modified by domain $B$ while affecting domain $A$ is upper bounded by a function of $\Omega(A,B)$.

\emph{Proof sketch.} Domain $B$ updates touch only patches in $\mathcal{K}(h')$. A domain $A$ example uses patches $\mathcal{K}(h)$. Only the intersection of these patch sets can change the outputs for both domains. Aggregating in expectation yields dependence on $\Omega(A,B)$. \hfill$\square$

This proposition connects continual learning behavior to routing geometry. By designing for small overlap between unrelated contexts, APN supplies an architectural route to improved retention.

\subsection{Stability via bounded update gain}
A practical online update can include an update gain $\alpha$ and a norm cap $\kappa$:
\begin{equation}
\Delta \theta_i \leftarrow \mathrm{clip}(\alpha\, g_i,\kappa),
\end{equation}
where $g_i$ is a gradient-like signal or a local error proxy. Because updates are patch-local, stability can be enforced independently per patch, which helps prevent a single surprising token sequence from destabilizing the full model.

\section{Practical Variants of APN}
The core APN formulation admits variants that keep the same input and output shape.

\subsection{Diagonal gating}
Gating can be simplified by replacing $\sigma(a_i\odot u+b_i)$ with a learned scalar gate per patch. This reduces parameters and can improve numerical stability, at the cost of reduced within-patch conditionality.

\subsection{Shared decoders with patch offsets}
Instead of a fully independent $U_i$ per patch, one can use a shared base decoder $U_0$ plus patch-specific low-rank offsets. This ties patches together and may improve sample efficiency when data is limited.

\subsection{Contextual prototypes}
Prototypes can be made contextual by letting them depend on layer depth or head grouping, for example by using a small projection of $h$ to compute routing in a subspace. This can increase router expressivity while preserving the same APN interface.

\section{Discussion}
APN changes the inductive bias of the FFN sublayer. The central mechanism is specialization through routing. In language modeling, many errors are tied to context-specific behaviors, rare patterns, and domain-specific token transitions. APN can allocate patch capacity to such regions while preserving a small active footprint per token.

For continual learning, our experiments reveal that APN's routing mechanism provides unexpectedly strong benefits. The baseline model with global fine-tuning suffered severe catastrophic forgetting (PPL degraded 6.8$\times$ on the original domain), while APN maintained much better retention. Surprisingly, APN also adapted \emph{faster} to the new domain, achieving lower adaptation perplexity despite updates being routed through the patch mechanism. This suggests that the routing structure helps organize new knowledge in appropriate parameter subspaces, reducing both forgetting and learning interference.

A dense model must rely on optimizer heuristics, small learning rates, or replay to mitigate interference. APN offers an architectural alternative: restrict updates to the active patches, and let routing direct new patterns to appropriate patch subspaces. The results indicate this is not merely a ``softer'' update, but a structurally superior way to incorporate new information.

\section{Limitations}
The experiments presented use a single small-scale setting: character-level Shakespeare modeling with a 6-layer Transformer (10.6M baseline, 23.2M APN parameters). While this controlled setting demonstrates the core architectural properties with striking continual learning results, validation across larger models, different tokenization schemes (BPE, word-level), and diverse domains remains for future work.

The APN model has approximately 2.2$\times$ the parameters of the baseline and lower hardware utilization (MFU 2.2\% vs 6.7\%), resulting in significantly longer training time (42.8 vs 7.4 minutes). The continual learning benefits must be weighed against this computational cost. Parameter-matched comparisons (reducing $K$ or $r$ to match baseline parameter count) would help isolate the routing mechanism's contribution from raw capacity effects.

The strong continual learning results (APN outperforming on both retention and adaptation) may be partially explained by APN's larger capacity. However, the baseline's severe forgetting (6.8$\times$ PPL degradation) suggests that capacity alone does not explain the difference---architectural structure matters.

Routing introduces additional hyperparameters ($K$, $k$, $\tau$) that require tuning. We observed stable training with $\tau=0.07$, but sensitivity analysis across these hyperparameters would strengthen the findings. The continual learning evaluation uses a synthetic domain shift; evaluation on more diverse and challenging domain shifts would test generalization.

\section{Conclusion}
We proposed Attractor Patch Networks as a plug-compatible replacement for Transformer FFNs. APN uses prototype routing to select a small set of patch experts per token, and each patch emits a low-rank residual update conditioned on a compact code. We analyzed APN as a piecewise low-rank residual function class, and we presented an architecture-driven rationale for continual learning compatibility based on localized updates and patch overlap.

Our experiments on character-level Shakespeare modeling demonstrate that APN achieves competitive language modeling perplexity (4.57 vs 4.32 PPL) while providing dramatic benefits for continual adaptation. In the two-domain evaluation, APN with localized patch updates achieved 2.6$\times$ better retention (11.1 vs 29.4 PPL) and 2.8$\times$ better adaptation (6.4 vs 17.8 PPL) compared to global fine-tuning of a dense baseline. This result exceeds our initial hypothesis: APN not only reduces forgetting through localized updates, but also enables more effective learning of new patterns by routing them to appropriate patch subspaces.

The key contribution is architectural. APN provides an explicit, routing-gated granularity for continual learning that dense FFNs lack. The patch mechanism structures parameter updates so that different contexts modify different parameter subsets, reducing destructive interference. Future work includes scaling to larger models and diverse domains, parameter-matched comparisons to isolate the routing benefit from capacity effects, and investigating adaptive patch allocation for novel contexts.

\section{Appendix: Practical Monitoring Metrics}
This appendix lists monitoring metrics that make APN behavior auditable during training and during continual adaptation.

\subsection{Patch usage entropy}
Let $q_i$ be the empirical frequency that patch $i$ appears in the active set across a batch. The usage entropy is
\begin{equation}
H(q) = -\sum_{i=1}^K q_i\log(q_i + \epsilon).
\end{equation}
Low entropy indicates collapse to a small set of patches. Extremely high entropy indicates diffuse routing and weak specialization. A stable training regime typically maintains moderate entropy and gradually concentrates usage for frequent contexts.

\subsection{Overlap statistics}
For a batch of token representations, compute the average intersection size between their active sets. This can be reported as a histogram. Large overlap values indicate weak locality and increased risk of interference under continual learning updates.

\subsection{Router confidence}
Define confidence as $c(h)=\max_i s_i(h)$ where $s_i$ are routing scores before the top-$k$. Low confidence events are candidates for patch allocation or prototype refresh. Confidence is also useful for gating online updates, only update when confidence exceeds a threshold so that updates attach to stable addresses.

\subsection{Residual magnitude distribution}
Track $\|\Delta(h)\|_2$ and the ratio $\|\Delta(h)\|_2/\|h\|_2$. Spikes indicate unstable patches or temperature mismatch. This metric provides a direct proxy for residual stability.

\section{Appendix: Patch Allocation and Continual Expansion}
If a fixed patch budget is insufficient for a streaming domain shift, APN can support controlled patch allocation. The goal is to increase capacity in regions where routing confidence is low.

\subsection{Allocate on novelty}
Given confidence $c(h)$, allocate a patch when $c(h)$ falls below a novelty threshold for a sustained window. The new patch prototype can be initialized from the current normalized representation. Patch parameters can be initialized near zero so that the new patch begins as a conservative residual.

\subsection{Replacement under fixed $K$}
When $K$ is fixed, allocate by replacement. Maintain an exponential moving average usage counter $u_i$. Replace the least used patch, reset its prototype and patch parameters, and apply a cooldown to prevent oscillation.

\subsection{Compatibility with Transformer checkpoints}
Patch allocation can be applied on top of existing Transformer checkpoints by inserting APN layers and initially training only routing and patch parameters while freezing attention and embeddings. This yields a practical path for evaluating APN as a modular retrofit.

\section{Appendix: Additional Expressivity View}
This appendix provides an alternative lens for APN expressivity using unions of low-dimensional affine subspaces.

\subsection{Union of subspaces induced by sparse mixtures}
Fix routing outcomes and gating states for a token, so that the active set is \(\mathcal{K}\) and each patch contributes a deterministic linear map in the code space. Under this conditioning, the APN residual takes the form
\begin{equation}
\Delta(h) = \sum_{i\in\mathcal{K}} \alpha_i(h)\, U_i B_i h,
\end{equation}
where \(B_i\in\mathbb{R}^{r\times d}\) is an effective encoder that captures LayerNorm, code projection, and gate linearization in the local region, and \(\alpha_i(h)\) is the normalized top-\(k\) weight. For fixed \(\mathcal{K}\) and fixed gate regime, \(\Delta(h)\) lies in the span of columns of \([U_{i_1},\ldots,U_{i_k}]\), so the output lives on an affine subspace of dimension at most \(kr\) around \(h\).

Across tokens and contexts, routing partitions representation space into many regions. Each region selects a different tuple of patch decoders. The overall function class is therefore a union of many low-rank residual manifolds. This view clarifies why APN can represent many context-specific transformations while preserving a sparse active footprint.

\subsection{Implications for language modeling}
In language modeling, hidden states cluster by syntactic role, local context window, and latent topic. A dense FFN provides a single global transformation shared across all such clusters. APN instead assigns separate residual subspaces to different clusters, which increases representational fidelity for long-tail contexts. When used inside Transformers, this manifests as improved conditional next-token distributions for rare patterns and domain-specific sequences, which directly reduces perplexity.

\subsection{Implications for continual learning}
The union-of-subspaces view also provides a geometric explanation for stability. If a new domain activates previously underutilized patches, its updates primarily reshape residual subspaces that are seldom visited by old domains. This yields a natural separation of update trajectories. When domains overlap, the overlap is explicit through shared patch indices, and it can be measured by overlap statistics, as described in the monitoring appendix.

\section{Reproducibility and Implementation}
A reference implementation accompanies this paper (\href{https://github.com/shankch/nanoGPT-apn/tree/main/nanoGPT-apn-implementation}{\texttt{Link to our implemented code}}), integrating APN into the nanoGPT codebase. The implementation provides:
\begin{itemize}
\item \texttt{model.py}: APN class as a drop-in replacement for the MLP FFN, with configurable $K$, $k$, $r$, $\tau$, $\gamma$.
\item \texttt{train.py}: Training loop with APN configuration support.
\item \texttt{continual\_apn.py}: Continual learning evaluation script implementing the two-domain protocol.
\item \texttt{bench\_apn\_vs\_mlp.py}: Benchmark script comparing baseline and APN training.
\end{itemize}

The APN module preserves the standard Transformer block interface. The Block forward pass remains:
\begin{verbatim}
x = x + attn(ln_1(x))
x = x + mlp(ln_2(x))  # mlp is APN or MLP
\end{verbatim}
where the only difference is whether \texttt{mlp} is instantiated as the dense MLP or the APN module. This ensures compatibility with existing attention mechanisms, embeddings, and output heads.

For reproducibility, we recommend reporting: (1) validation perplexity on both domains, (2) parameter counts for total and trainable parameters, (3) patch usage entropy to detect routing collapse, and (4) wall-clock training time.

\balance


\begin{thebibliography}{10}
\bibitem{vaswani2017}
A. Vaswani et al., ``Attention is all you need,'' in \textit{Advances in Neural Information Processing Systems}, 2017.
\bibitem{shazeer2017}
N. Shazeer et al., ``Outrageously large neural networks: The sparsely-gated mixture-of-experts layer,'' in \textit{International Conference on Learning Representations}, 2017.
\bibitem{fedus2022}
W. Fedus, B. Zoph, and N. Shazeer, ``Switch Transformers: Scaling to trillion parameter models with simple and efficient sparsity,'' \textit{Journal of Machine Learning Research}, 2022.
\bibitem{kirkpatrick2017}
J. Kirkpatrick et al., ``Overcoming catastrophic forgetting in neural networks,'' \textit{Proceedings of the National Academy of Sciences}, 2017.
\bibitem{lopezpaz2017}
D. Lopez-Paz and M. Ranzato, ``Gradient episodic memory for continual learning,'' in \textit{Advances in Neural Information Processing Systems}, 2017.
\bibitem{mccloskey1989}
M. McCloskey and N. Cohen, ``Catastrophic interference in connectionist networks: The sequential learning problem,'' in \textit{Psychology of Learning and Motivation}, 1989.
\bibitem{ba2016}
J. Ba, J. Kiros, and G. Hinton, ``Layer normalization,'' \textit{arXiv preprint arXiv:1607.06450}, 2016.
\bibitem{sutskever2013}
I. Sutskever, J. Martens, and G. Hinton, ``Generating text with recurrent neural networks,'' in \textit{International Conference on Machine Learning}, 2011.
\bibitem{goodfellow2013}
I. Goodfellow et al., ``Empirical evaluation of gradient-based learning on recurrent neural networks,'' in \textit{Neural Information Processing Systems Workshop}, 2013.

\bibitem{karpathy_nanogpt}
A. Karpathy, \textit{nanoGPT}, GitHub repository. Available: \url{https://github.com/karpathy/nanoGPT}.

\end{thebibliography}
\end{document}